\documentclass[conference]{IEEEtran}
\IEEEoverridecommandlockouts

\usepackage{cite}
\usepackage{amsmath,amssymb,amsfonts}
\usepackage{algorithmic}
\usepackage{graphicx}
\def\BibTeX{{\rm B\kern-.05em{\sc i\kern-.025em b}\kern-.08em
    T\kern-.1667em\lower.7ex\hbox{E}\kern-.125emX}}

\usepackage{booktabs} 
\usepackage{textcomp}
\usepackage{xcolor}
\usepackage{listings}
\usepackage{tabularx}

\usepackage{arabtex} 
\usepackage{utf8} 
\definecolor{codegreen}{rgb}{0,0.6,0}
\definecolor{codegray}{rgb}{0.5,0.5,0.5}
\definecolor{codepurple}{rgb}{0.58,0,0.82}
\definecolor{backcolour}{rgb}{0.95,0.95,0.92}
\definecolor{darkred}{rgb}{0.6, 0, 0}

\lstdefinestyle{mystyle}{
  backgroundcolor=\color{backcolour}, commentstyle=\color{codegreen},
  keywordstyle=\color{magenta},
  numberstyle=\tiny\color{codegray},
  stringstyle=\color{codepurple}, basicstyle=\ttfamily\footnotesize\color{darkred},
  breakatwhitespace=false,         
  breaklines=true,                 
  captionpos=b,                    
  keepspaces=true,                 
  numbers=left,                    
  numbersep=5 pt,  
  xleftmargin=1em,                 
  showspaces=false,                
  showstringspaces=false,
  showtabs=false,                  
  tabsize=2,
  moredelim=[is][\textcolor{blue}]{@}{@},
  moredelim=[is][\textcolor{codegreen}]{@@}{@@},
  moredelim=[is][\textcolor{black}]{@@@}{@@@},
  postbreak=\mbox{\textcolor{black}{$\hookrightarrow$}\space},
}
\begin{document}

\title{Prompt Engineering Techniques for Context-dependent Text-to-SQL in Arabic \thanks{* These authors contributed equally to this work.}
}

\author{\IEEEauthorblockN{1\textsuperscript{st} Saleh Almohaimeed *}
\IEEEauthorblockA{\textit{Department of Computer Science} \\
\textit{University of Central Florida}\\
Orlando, Florida, USA \\
sa247216@ucf.edu}
\and
\IEEEauthorblockN{2\textsuperscript{nd} May Alsofyani *}
\IEEEauthorblockA{\textit{Department of Computer Science} \\
\textit{University of Central Florida}\\
Orlando, Florida, USA \\
may\_sof@knights.ucf.edu}
\and
\IEEEauthorblockN{3\textsuperscript{rd} Saad Almohaimeed}
\IEEEauthorblockA{\textit{Department of Computer Science} \\
\textit{University of Central Florida}\\
Orlando, Florida, USA \\
sa583575@ucf.edu}
\and[\hfill\mbox{}\par\mbox{}\hfill]

\IEEEauthorblockN{4\textsuperscript{th} Mansour Al Ghanim}
\IEEEauthorblockA{\textit{Department of Computer Science} \\
\textit{University of Central Florida}\\
Orlando, Florida, USA \\
mansour.alghanim@ucf.edu}
\and
\IEEEauthorblockN{5\textsuperscript{th} Liqiang Wang}
\IEEEauthorblockA{\textit{Department of Computer Science} \\
\textit{University of Central Florida}\\
Orlando, Florida, USA \\
lwang@cs.ucf.edu}
}

\maketitle

\begin{abstract}
In recent years, the task of cross-domain, context-dependent text-to-SQL has received significant attention. Enables users with no prior knowledge of SQL to have a conversation with databases using natural language. However, most of the available datasets and research have been conducted in English, along with some work in Chinese. To this date, no effort has been made to address this task in the Arabic language. In this paper, we introduce Ar-SParC \footnote{Our dataset avaliable at github.com/Saleh-Almohaimeed/ar-sparc}, the first Arabic cross-domain, context-dependent text-to-SQL dataset. The dataset consists of 3,450 sequences of interrelated questions, each sequence containing an average of approximately three questions, which results in a total of 10225 questions along with their corresponding SQL queries. We conducted 40 experiments on the Ar-SParC dataset using two large language models, GPT-3.5-turbo and GPT-4.5-turbo, applying 10 different prompt engineering techniques, including four question representation methods and six in-context learning techniques. Furthermore, we developed a novel approach named GAT corrector, which enhanced the performance across all 40 experiments, yielding an average improvement of 1.9\% in execution accuracy (EX) and 1.9\% in interaction accuracy (IX) under zero-shot settings, and an average increase of 1.72\% EX and 0.92\% IX under in-context learning settings. Finally, we conducted an ablation study with two more experiments to explain why the GAT corrector outperformed the previous GAT verifier technique, particularly for the Arabic language.
\end{abstract}

\begin{IEEEkeywords}
Semantic Parsing, Text-to-SQL, Prompt Engineering
\end{IEEEkeywords}

\begin{figure}[t]
  \includegraphics[width=\columnwidth]{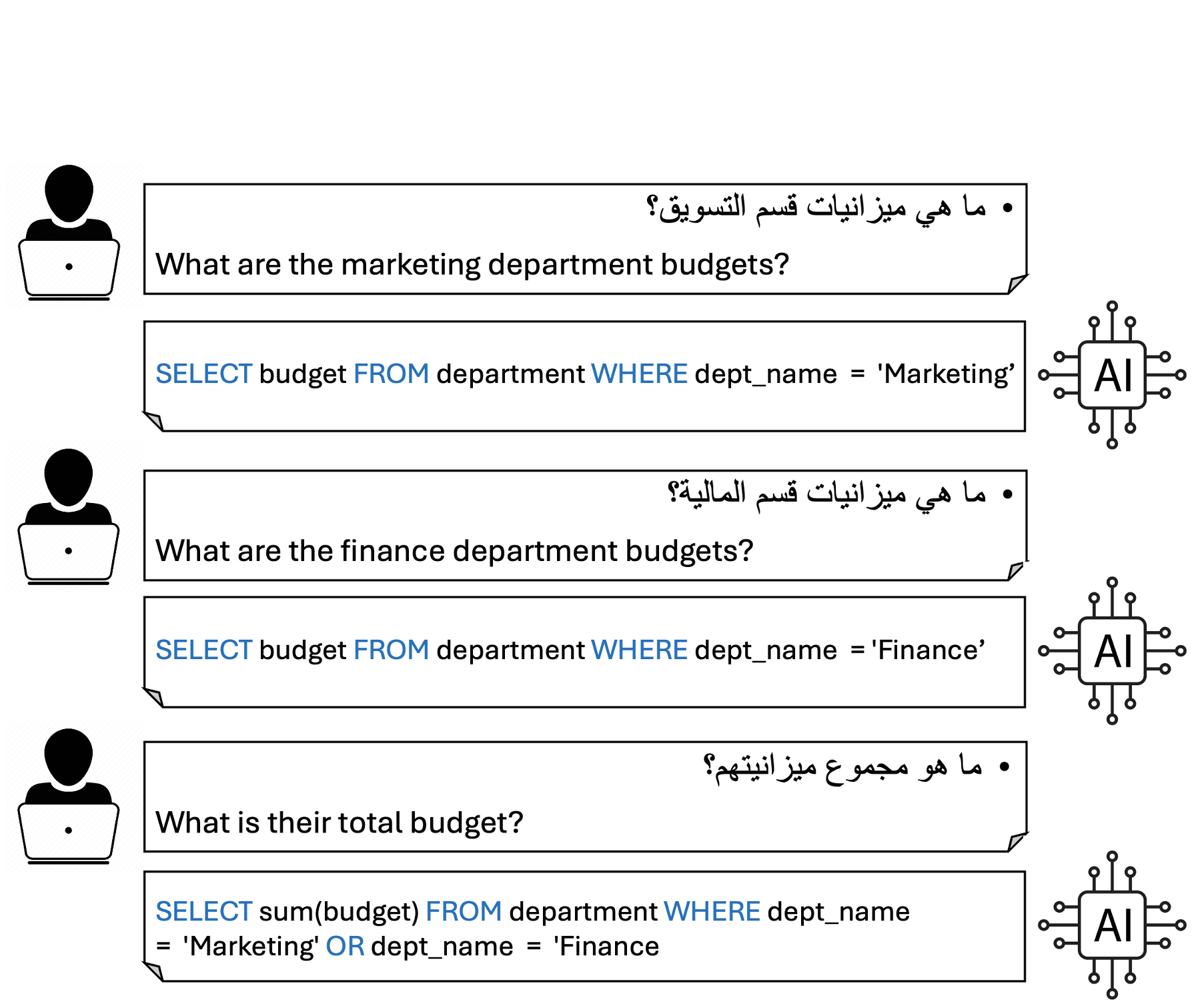}
  \caption{Examples of questions paired with their corresponding SQL queries in Ar-SParC. The first question is at the top, while the conversation progresses to the final question at the bottom.}
  \label{fig:conversation}
\end{figure}

\section{Introduction}
Accessing information from databases usually demands some basic knowledge of SQL, a language designed specifically for querying databases. For individuals who are unfamiliar with SQL, retrieving information from databases can be challenging. Therefore, the text-to-SQL task in the natural language processing (NLP) field changes this by enabling users to ask questions in natural language, and then SQL queries are automatically generated.

Currently, most research studies focus on single-question interactions between the user and the artificial intelligence (AI) model, in which the questions from the user are completely independent. Several datasets exist, including BIRD \cite{li2024can} in English, PAUQ \cite{bakshandaeva2022pauq} in Russian, and Ar-Spider \cite{almohaimeed2024ar} in Arabic. However, in real-life situations, users usually ask multiple dependent questions to obtain the information they need. This type of task is known as context-dependent text-to-SQL. There has been a significant amount of research on this task, along with various datasets released including ATIS \cite{price1990evaluation}, SParC \cite{yu2019sparc}, and CoSQL \cite{yu2019cosql} in English.

Despite these advancements, there is still a significant gap in addressing the needs of non-English languages, such as Arabic. With over 400 million Arabic speakers worldwide, the development of Arabic NLP resources is essential. Thus, we present Ar-SParC, a cross-domain context-dependent text-to-SQL dataset in Arabic. It consists of 3450 context-dependent question sequences, with an average of around three questions per sequence, resulting in a total of 10225 questions with corresponding SQL queries across 160 databases on 116 different domains. An example of Ar-SParC samples is presented in figure \ref{fig:conversation}, where users pose multiple dependent questions to get the desired final information. It is evident that the third question relies entirely on the response of the first and second questions. Consequently, if the model provides an incorrect answer to the first or second questions, it will likely result in erroneous responses to the third questions as well. This illustrates the inherent difficulty associated with context-dependent text-to-SQL conversions.

In the English SParC \cite{yu2019sparc} dataset, the best performance was achieved by GAT-SQL \cite{almohaimeed2024gat}, an advanced approach leveraging three distinct prompt engineering techniques: GAT representation, GAT reviser, and GAT verifier. These techniques were applied to prompt two large language models, GPT-3.5-turbo \cite{openai2024chatgpt} and GPT-4.5-turbo \cite{openai2023models}, in both zero-shot and in-context learning experiments.

In our empirical evaluation of the Ar-SParC dataset, we applied the same prompt engineering techniques as previous studies \cite{pourreza2024din} \cite{nan2023enhancing} \cite{chang2023prompt} \cite{openai2023sqltranslate} and introduced an enhancement to the GAT verifier \cite{almohaimeed2024gat}, which we named the GAT Corrector. This new technique is more efficient, both in terms of speed and cost and leads to an average performance improvement of 1.9\% in EX and 1.9\% in IX for zero-shot experiments, as well as 1.72\% in EX and 0.92\% in IX for in-context learning experiments. In addition, to show the effectiveness of the GAT Corrector, we performed an ablation study, demonstrating that it is better suited to the Arabic language than the GAT Verifier \cite{almohaimeed2024gat}.

The contributions of this paper are as follows: 

(1) we present Ar-SparC the first Arabic context-dependent cross-domain text-to-SQL dataset with 3450 sequences of questions across 160 databases on 116 different domains. 

(2) A total of 40 experiments were done with two large language models, GPT-3.5-Turbo \cite{openai2024chatgpt} and GPT-4.5-Turbo \cite{openai2023models}, in an analysis of different prompt engineering techniques, including four methods for question representation, and six methods for in-context learning selection techniques. 

(3) We propose the GAT Corrector prompt engineering technique, developed using GPT-3.5-Turbo \cite{openai2024chatgpt}, that has been fine-tuned on 500 samples created by our team. In both zero-shot and in-context learning techniques, the GAT Corrector significantly increases the performance achieving an average improvement of 1.9\% EX, 1.9\% IX in the zero-shot exepriment and 1.72\% EX, 0.92\% IX in the in-context learning experiment.

(4) We have also demonstrated the efficiency of GAT corrector for the Arabic language by comparing it with GAT verifier on a task other than text-to-SQL. There were only 3 mistakes by the GAT corrector out of 100, whereas 33 mistakes were made by the GAT verifier.
\section{Related Works}

Semantic parsing has been widely studied, which focuses on converting natural language sentences into formats that computers can understand, like SQL or Python, or mathematical expressions such as lambda calculus. Many datasets have been released in the past few years. In this section, we will look at the task from three perspectives. First, we will discuss datasets created for retrieving information from databases where the target format is not SQL. Second, we will focus on text-to-SQL datasets for context-independent tasks, where users ask independent questions, and each gets its own SQL query. Lastly, we will review recent work on context-dependent text-to-SQL, where users ask multiple dependent questions, as shown in Figure \ref{fig:conversation}.

There are several tasks that each contribute in a unique way to the first perspective, where the target representation is something other than SQL. For instance, the SCAN \cite{higgins2017scan} dataset translates natural language sentences into a sequence of actions. It is primarily designed to determine how well artificial intelligence models can generalize to actions that have not been specifically trained on. Another example is SIGMA \cite{almohaimeed2023sigma} dataset that uses Python as the target representation to retrieve information from databases rather than SQL. This allows it to not only retrieve information in its original format but also perform statistical analysis on the data before presenting it to the user. A third example is NL2Bash \cite{lin2018nl2bash}, which focuses on converting English sentences into Bash commands. The goal here is to allow users to carry out tasks like searching, retrieving, and manipulating files by writing their needs in plain English.

For the second perspective, where SQL is the target representation and the user's questions are completely independent, some very early attempts such as GroQuery \cite{zelle1996learning}, which is a small dataset built on a single domain. Later, more comprehensive datasets like Spider \cite{yu2018spider} and BIRD \cite{li2024can} were introduced. These datasets have been widely used in research, covering multiple databases across various domains, which helps models trained and tested on them to generalize better to new, unseen domains. Additionally, there have been efforts to create non-English datasets, many of which are translated versions of Spider \cite{yu2018spider} in different languages, such as Ar-Spider \cite{almohaimeed2024ar} in Arabic, CSpider \cite{min2019pilot} in Chinese, PAUQ \cite{bakshandaeva2022pauq} in Russian, and VSpider \cite{nguyen2020pilot} in Vietnamese.

The third perspective also uses SQL as the target representation, but the questions in this task are dependent on each other and necessitate context awareness for accurate translation. These context-dependent text-to-SQL task introduce new challenges because models should not only understand the structure of each query but also maintain the conversation history to effectively capture the user’s intent. Key datasets in this task include SParC \cite{yu2019sparc} and CoSQL \cite{yu2019cosql} in English, with efforts in Chinese such as Chase \cite{guo2021chase} and SeSQL \cite{huang2023sesql}, both of which focus on interactions where questions build on previous ones, which requires the model to remain aware of context and adjust its predictions accordingly.

Even though the aforementioned datasets have provided the foundation for multilingual and multidomain text-to-SQL task. There is still a significant gap in the availability of other multilingual datasets, especially for context-dependent task. This is where our proposed Arabic dataset for context-dependent text-to-SQL task comes into play. by the developing of Ar-SParC, this dataset will fill the void in Arabic language resources for conversational SQL parsing, enabling further research and development of models capable of understanding Arabic in database query contexts.

\section{Dataset}

\begin{table}[h]
\centering
\caption{A comparison between the English SParC dataset and the Arabic version Ar-SParC. \#Q denotes the questions, \#DB denotes the database, and \#Tables/DB represents the average number of tables per database.}
\label{tab:dataset}
\renewcommand{\arraystretch}{1.3}
\setlength{\tabcolsep}{6.5pt}
\begin{tabular}{cc|cccc}
\cline{3-6}
                                                      &                & \textbf{\#Q} & \textbf{\#Seq} & \textbf{\#DB} & \textbf{\#Tables/DB} \\ \hline
\multicolumn{1}{c|}{\textbf{English}}                 & \textbf{all}   & 12,726        & 4298           & 200           & 5.1                  \\ \hline
\multicolumn{1}{c|}{{\textbf{Arabic}}} & \textbf{all}   & 10,225         & 3,450          & 160           & 5.11                 \\
\multicolumn{1}{c|}{}                                 & \textbf{train} & 9,025         & 3,029          & 140           & 5.26                 \\
\multicolumn{1}{c|}{}                                 & \textbf{test}  & 1,200         & 421            & 20            & 4.05                 \\ \hline
\end{tabular}
\end{table}

Ar-SParC is an Arabic adaptation of the well-known English SParC \cite{yu2019sparc} dataset. It consists of two parts: training and development sets. Unlike the English version, we did not translate the test set, as the authors of SParC \cite{yu2019sparc} decided to keep it hidden for evaluation purposes. As detailed in table \ref{tab:dataset}, the training set consists of 3029 sequences, each containing an average of around three questions per sequence, leading to a total of 9025 questions. On the other hand, the development set includes 421 sequences, with an average of approximately three questions per sequence, resulting in a total of 1200 questions. Furthermore, the dataset includes 160 databases across 116 domains. To ensure that the models trained on Ar-SParC can generalize effectively to unseen domains, there is no overlap between the databases and domains in the training and development sets.

Regarding the construction of the Ar-SParC dataset, it was carried out by a team of five individuals. Initially, we followed a similar approach to that used in creating Ar-Spider \cite{almohaimeed2024ar}. We employed a large language model at the beginning of the translation process, using GPT-4.5-turbo \cite{openai2023models} to translate all the questions of the SParC dataset into Arabic. Previous study \cite{almohaimeed2024gat} found that this approach increases translation accuracy, reduces translation time by over 50\%, and allows a wide variety of Arabic words to be used. Then, after the automated translation, two native Arabic-speaking translators with expertise in translation services reviewed each question to ensure both the accuracy of the translation and that the question accurately reflects the types of questions typically asked by humans. Subsequently, three Arabic-speaking graduate students in computer science reviewed the original English questions, validated the accuracy of the Arabic translations after been post-edited, and ensured that the SQL queries accurately corresponded to the translated Arabic questions.

\section{Methodology}

In the methodology section, we will briefly summarize the previous prompt engineering techniques applied for text-to-SQL tasks. Then, we will present our own technique GAT Corrector, which has enhanced the performance over baselines. 

There are two aspects to consider in previous works, \textbf{question representation techniques}, which specify how the components of the prompt (such as the question, schema, etc.) will be arranged in the prompt, and \textbf{in-context learning techniques}, which incorporate examples from external sources, such as the training set, into the prompt to guide the LLM's behavior to generate the desired SQL query.


\subsection{Question Representation methods}

\begin{lstlisting}[style=mystyle, caption=Basic Prompt]
Table student, columns = [StuID, Fname, LName]
Table dorm, columns = [dormid, dorm_name]

@@How many students are there?@@
@ @
\end{lstlisting}

\begin{lstlisting}[style=mystyle, caption=Text Representation Prompt]
@@@Given the following database schema:@@@
Table student, columns = [StuID, Fname, LName]
Table dorm, columns = [dormid, dorm_name]

@@@Answer the following question:@@@ 
@@How many students are there?@@
\end{lstlisting}

\begin{lstlisting}[style=mystyle, caption=Code Representation Prompt]
@@@ /*Given the following database schema: */ @@@
CREATE TABLE student (
     StuID number PRIMARY KEY,
     Fname text,
     LName text,
);

CREATE TABLE dorm (
     dormid number PRIMARY KEY,
     dorm_name text,
     FOREIGN KEY (dormid) REFERENCES Dorm_amenity (dormid)
);
@ @
@@@ /*Answer the following question:@@@ @@How many students are there?@@ @@@*/@@@
\end{lstlisting}

\begin{lstlisting}[style=mystyle, caption=OpenAI Demostration Prompt]
@@@###Complete sqlite SQL query only and with no explanation@@@
@@@### SQLite SQL tables , with their properties:@@@
@@@#@@@
@@@#@@@ student(StuID, Fname, LName)
@@@#@@@ dorm(dormid, dorm_name)
@@@#@@@
@@@###@@@ @@How many students are there?@@
\end{lstlisting}

There have been several efforts to modify the arrangement of text-to-SQL prompts for maximum efficiency. The first attempt was by \cite{pourreza2024din}, who introduced the \textbf{Basic prompt (BSp)} shown in listing 1, where the prompt consists only of the target question and the database schema. Unfortunately, the lack of additional context makes this approach more challenging for LLMs. In response, a prompt known as a \textbf{Text representation prompt (TRp)} \cite{nan2023enhancing} was proposed, where additional information is added before the schema and target question, as demonstrated in listing 2. This helps LLMs better understand the structure of the prompt. 

Another technique is the \textbf{Code representation prompt (CRp)} \cite{chang2023prompt}, where the database schema is presented in the same format used for generating schema elements in SQL, as shown in listing 3. Additionally, any extra information in the prompt is included as comments encapsulated by two signs /* and */. An additional technique, implemented by OpenAI in their release of GPT-3 \cite{openai2024chatgpt}, involves a prompt structure known as the \textbf{OpenAI demostration prompt} \cite{openai2023sqltranslate}. In this format, any text is preceded by a pound sign (\#) or three pound signs (\#\#\#), as shown in listing 4. At the beginning of the prompt, a comment instructs, "Complete SQLite SQL query only and with no explanation," which helps the LLM focus solely on generating the SQL query without unnecessary details.  Finally, they have demonstrated in \cite{almohaimeed2024gat} that including samples of the cells values to the prompt will definitely increase performance. Because it eliminates the ambiguity caused by certain tables and columns in the database. Therefore, we will include a glance at the database values in our experiments of all previous question representation techniques.

\subsection{In-context learning methods}

Many studies in text-to-SQL propose different in-context learning prompt engineering techniques, which utilize samples from training sets to improve the comprehension of LLMs, example shown in listing 5. The primary distinction between these techniques lies in the mechanism used to select samples from the training set. The chosen samples typically include the database schema, the question, and the corresponding SQL query.

Previous research proposed six mechanisms for this process. The first is the \textbf{Random technique} \cite{gao2023text}, commonly used as a baseline in various studies, where samples are randomly selected from the training set and included in the prompt. The second mechanism is \textbf{Question Similarity Selection (QTSs)} \cite{liu2021makes}, which involves comparing the target question posed by the user with all questions in the training set. The most similar question, along with its SQL query and schema, is then included in the prompt. This similarity is calculated using the cosine similarity function, based on the pre-trained model "all-mpnet-base-v2" \cite{allmpnetbasev22023}.
The third technique is \textbf{Masked Question Similarity Selection (MQSs)} \cite{gao2023text}, which works similarly as QTSs. However, before calculating the cosine similarity, all table and column names in the user's target question are replaced with the token "\textless MSK\textgreater", eliminating potential negative effects caused by domain-specific information.

The fourth technique, known as \textbf{Query Similarity Selection (QRSs)} \cite{nan2023enhancing}, unlike previous ones. Instead of comparing the target question to each question in the training set, this technique gives the target question to a model fine-tuned on the training set to predict an SQL query, which will be treated as approximations of the desired SQL query. The predicted SQL query is then compared to all SQL queries in the training set using the Jaccard similarity technique. The most similar SQL queries, along with their questions and database schemas, are included in the prompt. The \textbf{DAIL Selection  (DAILs)} \cite{gao2023text} is the fifth technique, which is a combination of both MQSs \cite{gao2023text} and QRSs \cite{nan2023enhancing} techniques. It begins by ranking the training samples based on their question similarity to the target question, discarding those that fall below a certain similarity threshold. The remaining samples are then re-ranked using the QRSs \cite{nan2023enhancing} technique. This results in a final list of samples ranked according to a combination of both question and SQL query similarities.

The final technique called the \textbf{GAT reviser (GATr)} \cite{almohaimeed2024gat}, achieved a significantly higher performance compared to previous approaches on both the SParC \cite{yu2019sparc} and CoSQL \cite{yu2019cosql} datasets. While it shares similarities with the QRSs and DAILs methods, it takes a different approach. Instead of comparing the predicted SQL query from the fine-tuned model with all SQL queries in the training set, the technique assumes the predicted SQL query is correct. The LLM is then tasked with revising this predicted SQL query, determining its correctness. If the query is correct, the LLM rewrites it; if it is incorrect, the LLM modifies it and provides the corrected SQL query.

\begin{lstlisting}[style=mystyle, caption=Prompt using In-context learning selection metehod]
@@@Given the following database schema:@@@
${Database Schema}
@@@Answer the following question:@@@ 
@@How many City are there? @@
@SELECT count(*) FROM City @

@@@Given the following database schema:@@@
${Database Schema}
@@@Answer the following question:@@@ 
@@How many teams do we have? @@
@SELECT count(teamID) FROM Team @

@@@Given the following database schema:@@@
${Database Schema}
@@@Answer the following question:@@@ 
@@How many books available? @@
@SELECT count(*) FROM Book @


@@@Given the following database schema:@@@
Table student, columns = [StuID, Fname, LName]
Table dorm, columns = [dormid, dorm_name]

@@@Answer the following question:@@@ 
@@#How many students are there?@@
\end{lstlisting}

\subsection{GAT-Corrector}

\begin{figure*}[t]
  \centering
  \includegraphics[width=\textwidth]{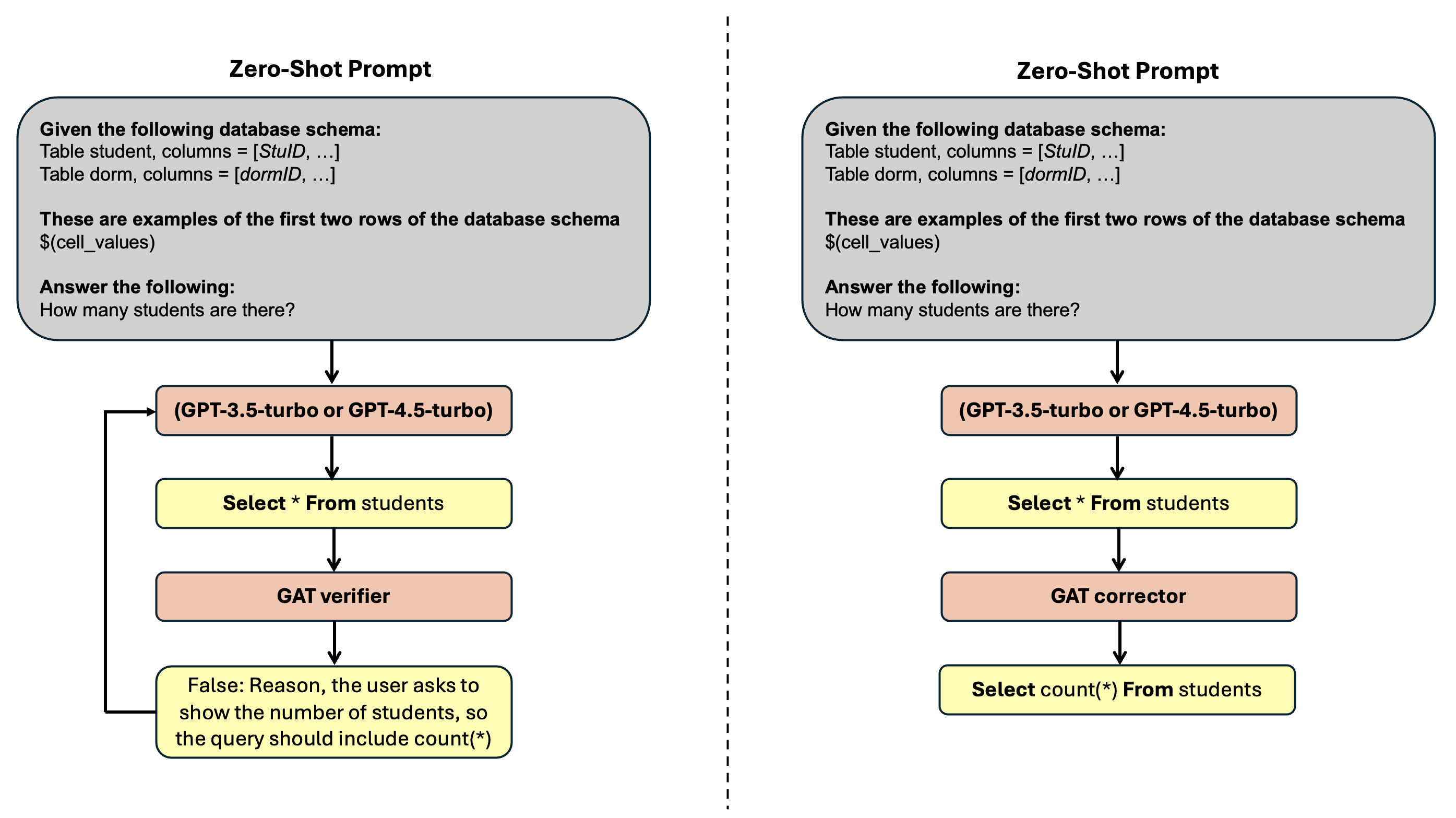}
  \caption{The process on the left shows how the GAT verifier works. First, the user's prompt is input into the LLM, which generates an SQL query. This query, along with the user question and schema, is then input into the GAT verifier to detect errors. The verifier outputs the result, which is then fed back to the LLM to correct any mistakes. On the other hand, The process on the right illustrates how the GAT corrector works, similar to the GAT verifier. However, instead of just detecting errors, the GAT corrector detects and corrects the errors at the same time.}
  \label{fig:GATCor}
\end{figure*}

In this chapter, we present GAT Corrector (GATc), a novel approach that builds upon and enhances the previously introduced GAT Verifier \cite{almohaimeed2024gat} (GATv) technique. Figure \ref{fig:GATCor} highlights the key differences between GAT Corrector and GAT Verifier \cite{almohaimeed2024gat}. Both techniques utilize the database schema, target question, and the predicted SQL query as input. For GAT Verifier, the output is a True or False result, along with an explanation. If the result is True, the system proceeds to the next question in the sequence. If the result is False, GAT Verifier \cite{almohaimeed2024gat} provides the LLM that generated the SQL query with an explanation of the error, and prompting it to generate a new revised SQL query.

This technique has three main limitations. First, although it performs well on the English SParC \cite{yu2019sparc} dataset, it faces challenges with the Arabic Ar-SParC dataset, often misclassifying incorrect SQL queries as correct or vice versa. Second, when the GAT Verifier returns a False response, it instructs the previous LLM to regenerate a new SQL query, leading to increased costs, especially with models like GPT-3.5-turbo \cite{openai2024chatgpt} and GPT-4.5-turbo \cite{openai2023models}. Third, the process of regenerating the SQL query introduces additional time overhead. While this may not be an issue for a single query, it becomes a significant challenge when handling thousands of queries.

To overcome the previously mentioned limitations, we introduce the GAT Corrector. Like GAT Verifier \cite{almohaimeed2024gat}, it takes the same inputs (database schema, target question, and predicted SQL query). However, instead of simply determining whether the SQL query is correct or false, GAT Corrector generates the same previous SQL query if it was correct and generatess a new SQL query if the previous is incorrect. In other words, GAT corrector detected and fix the errors at the same time. By eliminating the need to request another LLM to generate a new SQL query, this approach saves both time and money. In addition, it performs better with the Arabic Ar-SParC dataset. To understand why GAT Corrector succeeds with Arabic data while GAT Verifier does not, refer to the ablation study in section 5.4.

\begin{table*}[h]
\centering
\caption{Zero-shot experiments for different question representation techniques}
\label{tab:ZS no GATc}
\begin{tabularx}{\textwidth}{l *{5}{X}}
\toprule

 & \multicolumn{2}{c}{\textbf{GPT-3.5-turbo}} & \multicolumn{2}{c}{\textbf{GPT-4.5-turbo}} \\ 
 
\cmidrule(r){2-3} \cmidrule(lr){4-5}

\textbf{Question Representation}                                 & \textbf{EX} & \textbf{IX} & \textbf{EX} & \textbf{IX} \\ 
\midrule

\textbf{BSp}    & 56.6        & 35.6        & 64.5        & 42.8        \\

\textbf{CRp}     & 57.9        & 37.5        & 65.3& 43.2        \\

\textbf{ODp}   & \textbf{58.6}  & \textbf{37.8}& \textbf{65.6}    & \textbf{44.2}\\

\textbf{TRp}     & 57.7          & 37.3        & 64.9        & 43        \\ 

\bottomrule
\end{tabularx}
\end{table*}

\begin{table*}[]
\centering
\caption{In-Context-Learning experiments with OpenAI demostration technique}
\label{tab:ICL no GATc}
\begin{tabularx}{\textwidth}{l *{5}{X}}
\toprule

 & \multicolumn{2}{c}{\textbf{GPT-3.5-turbo}} & \multicolumn{2}{c}{\textbf{GPT-4.5-turbo}} \\ 
 
\cmidrule(r){2-3} \cmidrule(lr){4-5}

\textbf{Question Representation}                                 & \textbf{EX} & \textbf{IX} & \textbf{EX} & \textbf{IX} \\ 
\midrule

\textbf{DAILs}    & 59.3        & 39.7        & 65.2        & 44.2        \\

\textbf{MQSs}     & 57.5        & 36.8        & 65.2 & 43.9       \\

\textbf{QRSs}   & 60.7  & 41.1& 68.2    & 47.7\\

\textbf{QTSs}     & 57.3          & 37.1       & 65       & 43        \\ 

\textbf{Random}     & 56.6          & 35.6        & 63.2        & 42.5        \\

\textbf{GAT Reviser}     & \textbf{71.3}           & \textbf{50.1}        & \textbf{71.5 }       & \textbf{52.3}        \\

\bottomrule
\end{tabularx}
\end{table*}

\begin{lstlisting}[style=mystyle, caption=Example of sample used to fine-tune GAT Corrector]
"messages": [
{
    "role": "system",
    "content": "You are a helpful assistant that check if SQL query are correctly representing user Question."
},
{
    "role": "user",
    "content": 
       @@@${Schema}@@@ 
       "How many students are there?"
       @$SELECT * FROM student@
},
{
    "role": "assistant",
    "content": "SELECT count(*) FROM student"
}
]
\end{lstlisting}

\textbf{Implementation of GAT corrector},
We have first developed a dataset consisting of 500 samples, which we plan to use for fine-tuning an LLM, specifically GPT-3.5-turbo \cite{openai2024chatgpt}. Out of the 500 samples, 250 contain SQL queries that correctly represent the user’s question, while the other 250 do not. We ensured that the dataset includes a wide range of possible mistakes that an LLM might make when generating SQL queries, such as missing columns in the SELECT clause or ordering results in ascending order when they should be in descending order, among other mistakes.

Each sample is structured with three distinct roles, as shown in Listing 6. The first role is the system, where we define the main task for the LLM. In our case, it is: "You are a helpful assistant that checks if SQL queries correctly represent the user’s question." The second role is the user, which contains the actual prompt typically used to prompt LLM. In our case, it includes the user’s question, the SQL query, and the database schema. The third role is the assistant, which provides the expected response we aim to achieve when prompting our fine-tuned model.

\section{Experiments}

\begin{table*}[]
\centering
\caption{GAT Corrector experiments under zero-shot settings}
\label{tab:GATcZS}
\begin{tabularx}{\textwidth}{l *{5}{X}}
\toprule

 & \multicolumn{2}{c}{\textbf{GPT-3.5-turbo}} & \multicolumn{2}{c}{\textbf{GPT-4.5-turbo}} \\ 
 
\cmidrule(r){2-3} \cmidrule(lr){4-5}

\textbf{Question Representation}                                 & \textbf{EX} & \textbf{IX} & \textbf{EX} & \textbf{IX} \\ 
\midrule

\textbf{BSp}    & 59.4        & 38.2        & 66.5        & 44.2        \\

\textbf{CRp}     & 60.3        & 39.4        & \textbf{67.2}& \textbf{45.8}        \\

\textbf{ODp}   & \textbf{60.6}  & \textbf{39.8}& 67.1    & 44.7\\

\textbf{TRp}     & 59.4          & 38.2        & 66.6        & 45.1        \\ 

\bottomrule
\end{tabularx}
\end{table*}

\begin{table*}[]
\centering
\caption{GAT Corrector experiments under in-context learning settings}
\label{tab:GATcICL}
\begin{tabularx}{\textwidth}{l *{5}{X}}
\toprule

 & \multicolumn{2}{c}{\textbf{GPT-3.5-turbo}} & \multicolumn{2}{c}{\textbf{GPT-4.5-turbo}} \\ 
 
\cmidrule(r){2-3} \cmidrule(lr){4-5}

\textbf{Question Representation}                                 & \textbf{EX} & \textbf{IX} & \textbf{EX} & \textbf{IX} \\ 
\midrule

\textbf{DAILs}    & 60.1        & 40.1        & 65.9        & 44.2        \\

\textbf{MQSs}     & 59.8        & 38.7        & 66.6 & 44.2       \\

\textbf{QRSs}   & 62.3  & 42.3& 69.3    & 48.7\\

\textbf{QTSs}     & 59.6          & 38       & 65.9       & 43.2        \\ 

\textbf{Random}     & 58        & 36.3       & 66.2       & 43.9       \\

\textbf{GAT Reviser}     & \textbf{71.8}           & \textbf{50.1}        & \textbf{72.5 }       & \textbf{52.9}        \\

\bottomrule
\end{tabularx}
\end{table*}

In this section, we will discuss the most popular metrics for evaluating our experiments, and we will then present the results of our 40 experiments across four types of analysis. First, we will highlight eight experiments that explore different question representation techniques under zero-shot settings, where the prompt contains no external information, such as training examples. Next, we will showcase 12 experiments under few-shot settings (in-context learning), where external information is included in the prompt to improve the model's understanding. Following that, we will demonstrate how our new technique, GAT Corrector, enhances performance in both zero-shot and in-context learning experiments.  Finally, two experiments will be conducted to explain why GAT corrector is superior to GAT verifier \cite{almohaimeed2024gat} when dealing with Arabic.

\subsubsection{Evaluation Metrics}

Two commonly used evaluation metrics in text-to-SQL tasks are Execution Accuracy (EX) and Interaction Accuracy (IX). Execution Accuracy indicates the percentage of questions answered correctly by the model. Interaction Accuracy, on the other hand, measures how accurately the model answers sets of interrelated questions in one sequence.

\subsection{Zero-shot experiments}

These experiments focus on zero-shot prompting, where the schema and the user's question are the only outsource elements included in the prompt. As shown in Table \ref{tab:ZS no GATc}, OpenAI demonstration (ODp) \cite{openai2023sqltranslate} prompt produces the best performance in terms of execution and interaction accuracy. Most likely, this is because this representation technique ensures the LLM’s response contains no additional information, focusing only on generating the desired SQL query, without any unnecessary information or explanations.

Another noteworthy finding is that the Code representation prompt (CRp) \cite{chang2023prompt} ranked second in both the GPT-3.5-turbo \cite{openai2024chatgpt} and GPT-4.5-turbo \cite{openai2023models} experiments. This could be because the schema's tables and columns are written as SQL code, which implies to the model that the task involves SQL queries. Consequently, the LLMs more easily understand the database structure, as both GPT-3.5-turbo and GPT-4.5-turbo possess extensive knowledge of SQL programming.

An intriguing insight is the significant difference in Arabic language comprehension between GPT-3.5-turbo and GPT-4.5-turbo. Using the same question representation techniques, the results show a gap of approximately 7.38\% in execution accuracy and 6.25\% in interaction accuracy, which was not seen in the English SParC experiments \cite{yu2019sparc}. In the English dataset, for example, the text representation prompt (TRp) \cite{nan2023enhancing} achieved 65.3\% EX and 44.2\% IX on GPT-3.5-turbo, compared to 65.8\% EX and 44.2\% IX on GPT-4.5-turbo, indicating that the performance gap was significantly smaller.

\subsection{In-Context learning experiments}

The table \ref{tab:ICL no GATc} shows the results of in-Context Learning experiments, in which the LLM model generates the desired SQL query based on a few training samples included in the prompt as prior knowledge. As the ODp \cite{openai2023sqltranslate} achieves the best results under zero-shot settings, it has been used as the default question representation technique for all in-Context learning experiments. The experiments also reveal significant insights into the effectiveness of these techniques, with significant performance obtained when applying the GAT Reviser \cite{almohaimeed2024gat}. In addition, across the table, GPT-4.5-turbo outperforms GPT-3.5-turbo, which reemphasizes that GPT-4.5-turbo is better than GPT-3.5-turbo when dealing with Arabic. Moreover, we limited the in-context learning examples included in the prompt to 3 shots. We avoided using more samples due to the GPT-3.5-Turbo \cite{openai2024chatgpt} maximum input capacity of 4096 tokens. To ensure fair comparison between GPT-3.5-Turbo and GPT-4.5-Turbo \cite{openai2023models}, the token limit for GPT-4.5-Turbo was also restricted to 4096 tokens.

Furthermore, the GAT Reviser \cite{almohaimeed2024gat} technique yields the highest performance for both models, with GPT-4.5-turbo achieving 71.5\% EX and 52.3\% IX, while GPT-3.5-turbo recording 71.3\% EX and 50.1\% IX. This suggests that the GAT Reviser is ideally suited to enhance both execution and interaction accuracy, outperforming all other techniques. Random in-context learning technique, on the other hand, consistently result in lower accuracy scores, indicating that contextually-aware or structured approaches (like GAT Reviser) are necessary for maximizing model efficiency in this task.

The success of QRSs \cite{nan2023enhancing}, achieving 60.7\% EX and 41.1\% IX on GPT-3.5-turbo and 68.2\% EX and 47.7\% IX on GPT-4.5-turbo, can be attributed to its approach of encoding the predicted SQL queries into binary syntax vectors and applying the Jaccard similarity technique to find the most similar matching SQL queries from the training set. This method enhances the accuracy of SQL predictions by aligning the LLM output to actual examples of SQL syntax, reducing errors caused by natural language ambiguity. Unlike QTSs \cite{liu2021makes} and MQSs \cite{gao2023text}, which emphasize searching for similar candidate questions, QRSs \cite{nan2023enhancing} focus on the predicted SQL queries, making use of both structural and semantic similarities.

\subsection{GAT Corrector}

In tables \ref{tab:GATcZS} and \ref{tab:GATcICL}, the performance of the GAT Corrector is presented for all zero-shot and in-context learning experiments. For zero-shot experiments, the GAT Corrector enhances performance by an average of 1.9 \% EX and 1.9 \% IX across all 8 experiments. Moreover, for in-context learning experiments, it also enhances performance by an average of 1.72 \% EX and 0.92 \% IX across 12 experiments. In addition, the OpenAI demonstration prompt (ODp) \cite{openai2023sqltranslate} continues to perform the best, even with the GAT Corrector applied in GPT-3.5-turbo \cite{openai2024chatgpt}. However, this is not the case when using GPT-4.5-turbo \cite{openai2023models}, where ODp ranks second in execution accuracy after the Code Representation prompt (CRp) \cite{chang2023prompt}. Morover, ODp ranks third, in terms of interaction accuracy, behind both CRp and TRp.

Depending on the experiment, the GAT corrector can sometimes increase the EX metric more than the IX metric, or the opposite can happen. For instance, in zero-shot experiments using the GAT corrector, specifically with the Text representation Prompt (TRp) \cite{nan2023enhancing}, the EX increased by 1.7\% compared to not using the GAT corrector, while the IX increased by 0.9\%. Another example is in zero-shot experiments with the ODp prompt, where the EX increased by 1.4\%, and the IX increased by 2.1\%.

The reason for this difference is that the number of individual questions in the dataset is approximately three times the number of sequences. For example, in one sequence, the model could correctly answer two out of three questions, and when the GAT corrector is applied, it corrects the third question. This results in only a 0.083\% increase in EX, but a 0.23\% increase in IX. Consequently, if the GAT corrector mostly answers questions that make the entire sequence correct, the IX will outperform the EX metric.

GAT Corrector showed the largest improvement across all 20 experiments, including both zero-shot and in-context learning, with the BSp prompt in zero-shot settings, where it showed an increase of 2.8\% in EX and 2.6\% in IX. GAT Corrector success can be credited to its ability to effectively correct SQL queries in a single step without requiring further input from another LLM. By addressing both error detection and correction in one go, it reduces the chances of errors and cuts down on the need for multiple interactions with the LLMs. 

\subsection{Ablation Study}

\begin{table}[h]
\caption{A comparison of classification performance between GAT Verifier and GAT Corrector techniques in detecting if two sentences in two languages have the same meaning or not.}
\label{tab:my-GATC VS GATV}
\renewcommand{\arraystretch}{1.2}
\setlength{\tabcolsep}{20 pt}
\centering
\begin{tabular}{c|cc}
\hline
\textbf{Technique}  & \textbf{Correct} & \textbf{False} \\ \hline
\textbf{GAT Verifier}  & 67 / 100 & 33 / 100 \\ 
\textbf{GAT Corrector} & 97 / 100 & 3 / 100 \\ \hline
\end{tabular}
\end{table}

In both GAT verifier \cite{almohaimeed2024gat} and GAT corrector, the inputs are the database schema, target question, and SQL query, with the primary goal being to detect errors. However, the GAT corrector is also responsible for fixing any errors identified. Our ablation study reveals that the GAT corrector outperforms the GAT verifier, even though both perform essentially the same task, with the GAT corrector having additional complexity. Further, although the GAT verifier demonstrated strong results in experiments with the English SParC \cite{yu2019sparc} dataset, it did not perform as well with the Arabic Ar-SParC dataset. We hypothesize that fine-tuning the model only to classify queries as correct or false does not provide the LLM with sufficient information to detect errors. In contrast, fine-tuning the model to fix the problem by generating the correct SQL query in the event of an error detected, may provide more information and will help the LLM to be more effective.

To test this hypothesis, we used the GAT verifier \cite{almohaimeed2024gat} and GAT corrector for a task other than text-to-SQL, which is to determine whether two sentences in different languages have the same meaning. It would support our hypothesis if the GAT corrector once again outperforms the GAT verifier. Two prompts were designed: the first acts as GAT verifier, where it checks if an Arabic sentence and its English counterpart have the same meaning. In the case of similar meanings, the outputs will be "true", whereas, in the case of different meanings, the outputs will be "false" with an explanation. The second prompt, acting as the GAT corrector, has the same goal. However, if the meanings are the same, it will simply rewrite the Arabic question without changes, but if the meanings differ, it will modify and write the Arabic question so it matches the English meaning.

In our experiment, we selected the first 100 questions from the SParC English dataset, and the corresponding questions from the Ar-SParC Arabic dataset, and manually examined each result. According to table \ref{tab:my-GATC VS GATV}, the GAT verifier technique performs poorly in Arabic, with 33 out of 100 examples displaying incorrect results. In some cases, the explanations were even conflicting, such as in one instance where the explanation incorrectly stated, "The Arabic sentence is asking about the number of credits offered by each department, while the English sentence is asking about the number of credits offered by each department," which was clearly an error. On the other hand, the GAT corrector technique performed much better, with only 3 out of 100 examples flagged as having different meanings. Consequently, even if a model performs well in English, it may not necessarily perform well in other languages, emphasizing the importance of supporting low-resource languages.

\section{Conclusion}

This paper introduces Ar-SParC, a cross-domain, context-dependent text-to-SQL dataset. All questions were verified by two professional translators to ensure correctness and that they accurately reflect natural human questions, followed by three graduate students of computer science to ensure alignment with the corresponding SQL queries. On Ar-SParC, 40 experiments were conducted using various prompt engineering techniques, including question representation and in-context learning techniques. We also presented a novel method, the GAT Corrector, which significantly improved performance in all experiments, demonstrating its usefulness for Arabic. The performance of zero-shot experiments was increased by an average of 1.9\% EX and 1.9\% IX. In addition, it improved in-context learning experiments by an average of 1.72\% EX and 0.92\% IX. We conclude by explaining why the GAT Corrector outperformed the GAT Verifier and and emphasizes the need to support low-resource languages.

\end{document}